\documentclass[10pt,twocolumn,letterpaper]{article}
\usepackage{cvpr}              % To produce the CAMERA-READY version
\definecolor{cvprblue}{rgb}{0.21,0.49,0.74}
\usepackage[pagebackref,breaklinks,colorlinks,allcolors=cvprblue]{hyperref}
\usepackage{multirow}  % Add this package for multirow support
\usepackage{booktabs}  % For nicer tables
\usepackage{rotating}
\usepackage{amssymb}
\usepackage{comment}
\usepackage{bbm}
\usepackage[dvipsnames]{xcolor}

\def\paperID{8472} % *** Enter the Paper ID here
\def\confName{CVPR}
\def\confYear{2025}

\title{UniGraspTransformer: Simplified Policy Distillation for \\ Scalable Dexterous Robotic Grasping
\vspace{-2mm}}

\author{
Wenbo Wang$^{2}$ \quad Fangyun Wei$^{1}$\thanks{Corresponding author.} \quad Lei Zhou$^{3}$ \quad Xi Chen$^{1}$ \quad Lin Luo$^{1}$ \quad Xiaohan Yi$^{1}$ \\
Yizhong Zhang$^{1}$ \quad Yaobo Liang$^{1}$ \quad Chang Xu$^{2}$ \quad Yan Lu$^{1}$ \quad Jiaolong Yang$^{1}$ \quad Baining Guo$^{1}$\\ \\
$^1$Microsoft Research Asia \quad $^2$University of Sydney \quad $^3$National University of Singapore \\
\texttt{\small\{fawe,xichen6,liluo,yixiaohan,yizzhan,yalia,yanlu,jiaoyan,bainguo\}@microsoft.com}\\
\texttt{\small aaronwenbwang@gmail.com} \quad \texttt{\small leizhou@u.nus.edu}  \quad \texttt{\small c.xu@sydney.edu.au}
\vspace{-2mm}
}

\begin{document}
\maketitle
\begin{abstract}
We introduce UniGraspTransformer, a universal Transformer-based network for dexterous robotic grasping that simplifies training while enhancing scalability and performance. Unlike prior methods such as UniDexGrasp++, which require complex, multi-step training pipelines, UniGraspTransformer follows a streamlined process: first, dedicated policy networks are trained for individual objects using reinforcement learning to generate successful grasp trajectories; then, these trajectories are distilled into a single, universal network. Our approach enables UniGraspTransformer to scale effectively, incorporating up to 12 self-attention blocks for handling thousands of objects with diverse poses. Additionally, it generalizes well to both idealized and real-world inputs, evaluated in state-based and vision-based settings. Notably, UniGraspTransformer generates a broader range of grasping poses for objects in various shapes and orientations, resulting in more diverse grasp strategies. Experimental results demonstrate significant improvements over state-of-the-art, UniDexGrasp++, across various object categories, achieving success rate gains of 3.5\%, 7.7\%, and 10.1\% on seen objects, unseen objects within seen categories, and completely unseen objects, respectively, in the vision-based setting. 
Project page: \href{https://dexhand.github.io/UniGraspTransformer/}{https://dexhand.github.io/UniGraspTransformer/}.
\vspace{-2mm}
\end{abstract}    
\section{Introduction}
\label{sec:intro}

Dexterous robotic grasping~\cite{DexGraspNet, UniDexGrasp, UniDexGrasp++, dexterous_grasping_0, dexterous_grasping_1} remains a formidable challenge in the field of robotics, especially when dealing with objects that exhibit a wide variety of shapes, sizes, and physical properties. Dexterous hands~\cite{ShadowHand, InspireHand}, with their multiple degrees of freedom and complex control requirements, present unique difficulties in manipulation tasks. While methods such as UniDexGrasp++~\cite{UniDexGrasp++} have made notable progress in this area, they encounter significant performance degradation when a single network is tasked with a large and diverse set of objects. Additionally, UniDexGrasp++~\cite{UniDexGrasp++} employs a multifaceted training process, including policy learning, geometry-aware clustering, curriculum learning, and policy distillation, which complicates scaling and reduces efficiency.

\begin{figure}[!t]
    \centering
    \includegraphics[width=0.99\linewidth]{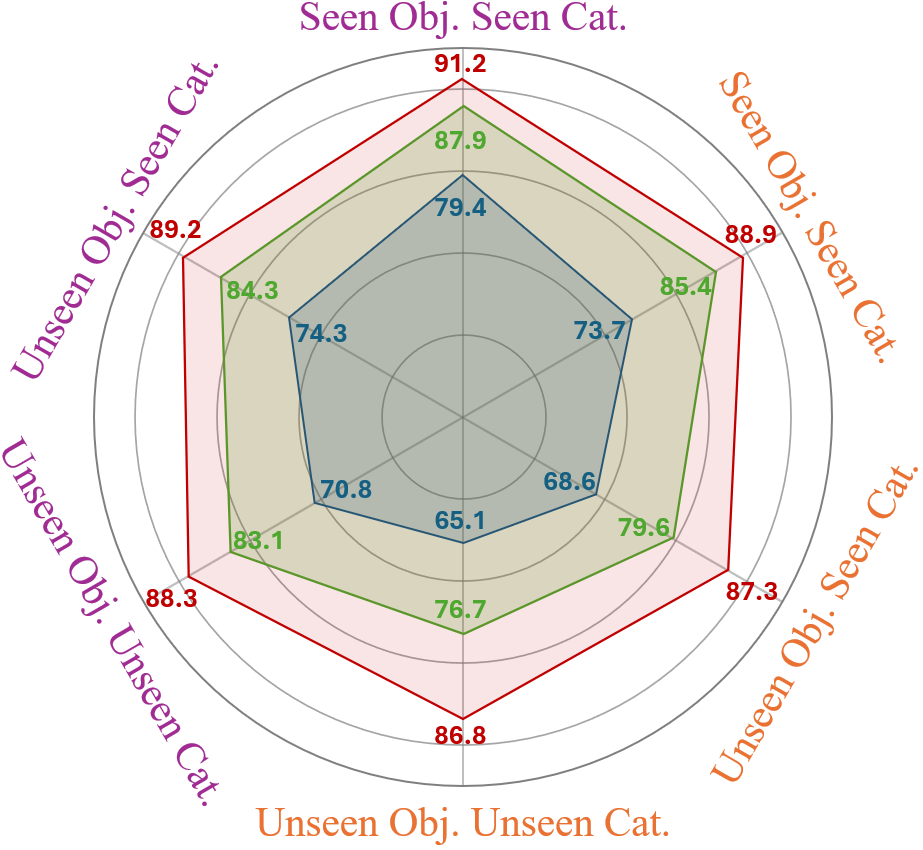}
    \vspace{-2mm}
    \caption{Performance comparison among \textcolor{MidnightBlue}{UniDexGrasp}~\cite{UniDexGrasp}, \textcolor{ForestGreen}{UniDexGrasp++}~\cite{UniDexGrasp++} and \textcolor{Maroon}{our UniGraspTransformer}, across \textcolor{Purple}{state-based} and \textcolor{Bittersweet}{vision-based} settings. For each setting, success rates are evaluated on seen objects, unseen objects within seen categories, and entirely unseen objects from unseen categories.}
    \vspace{-6mm}
    \label{fig:teaser}
\end{figure}

In this work, we simplify the training process of a universal network capable of handling thousands of objects while simultaneously improving both performance and generalizability. The workflow we propose is straightforward: 1) For each object in the training set, we begin by training a \textit{dedicated policy network} using reinforcement learning, guided by carefully crafted reward functions that enable the robot to master object-specific grasping strategies; 2) Next, these well-trained policy networks are used to generate millions of successful grasp trajectories; 3) Finally, we train a universal Transformer-based network, namely \textit{UniGraspTransformer}, in a supervised manner on this extensive trajectory set, allowing the network to generalize effectively to both the objects seen during training and new, unseen objects. Our architecture offers four key advantages:

\begin{itemize}
    \item \textbf{Simplicity.} We directly distill all dedicated reinforcement learning policies into a universal network in an offline style, without utilizing any extra techniques like network regularization or progressive distillation~\cite{UniDexGrasp++, GSL, progressive_distillation_0}.
    
    \item \textbf{Scalability.} Larger grasping networks generally demonstrate the ability to handle a broader range of objects and exhibit greater robustness to variations in shape and size. Our approach, which leverages offline distillation, allows the final network, UniGraspTransformer, to be designed at a larger scale, accommodating up to 12 self-attention blocks~\cite{Attention}. This provides significant flexibility and capacity compared to traditional online distillation methods~\cite{UniDexGrasp, UniDexGrasp++}, which often rely on smaller MLP networks to ensure convergence but limit scalability. Additionally, our dedicated policy networks are intentionally lightweight, as each only needs to handle a single object, ensuring efficiency without compromising performance.
    
    \item \textbf{Flexibility.} Each dedicated policy network is trained in a controlled, idealized environment where the full state of the system, including object representations (e.g., complete point clouds), dexterous hand states (e.g., finger-joint angles), and their interactions (e.g., hand-object distance), is fully observable and precisely accurate. Our architecture enables the distillation of knowledge from this ideal setting to more practical, real-world environments where some observations may be incomplete or unreliable~\cite{UniDexGrasp, UniDexGrasp++, vision_grasp_0, vision_grasp_1, vision_grasp_2, vision_grasp_3}. For instance, object point clouds might be noisy, or measurements of object poses may be imprecise. The primary role of these dedicated policy networks is to generate diverse, successful grasping trajectories across a wide range of objects. During the distillation process, these grasp trajectories serve as annotated data, enabling us to train our UniGraspTransformer model using \textit{realistic inputs} (e.g., noisy object point clouds and estimated object poses) to predict action sequences that closely mimic the successful grasp trajectories from the ideal setting.
    
    \item \textbf{Diversity.} In addition to being capable of grasping thousands of distinct objects, our larger universal network, coupled with the offline distillation strategy, demonstrates the ability to generate a broader range of grasping poses for objects presented in various orientations. This marks a significant improvement over prior methods, such as UniDexGrasp++~\cite{UniDexGrasp++}, which tend to produce repetitive, monotonous grasping poses across different objects. 
\end{itemize}

In our experiments, our proposed approach demonstrates substantial improvements over the previous state-of-the-art, UniDexGrasp++~\cite{UniDexGrasp++}, across a range of evaluation settings. Specifically, we evaluate our method in both the state-based setting—where object observations and dexterous hand states are perfectly accurate, as provided by a simulator—and the vision-based setting, where object point clouds are derived from multi-view reconstruction. Our approach consistently outperforms UniDexGrasp++~\cite{UniDexGrasp++} across multiple object types, including seen objects, unseen objects within seen categories, and entirely unseen objects from unseen categories, as illustrated in Figure~\ref{fig:teaser}. For instance, our method achieves performance gains of 3.5\%, 7.7\%, and 10.1\% over UniDexGrasp++~\cite{UniDexGrasp++} on seen objects, unseen objects within seen categories, and entirely unseen objects from unseen categories, under the vision-based setting.

\section{Related Works}
\label{sec:related_works}

\noindent\textbf{Robotic Grasping.} Robotic grasping~\cite{robot_grasping_0, robot_grasping_1, robot_grasping_2, robot_grasping_3} has been a longstanding research in robotics and computer vision, aiming to enable robots to interact with objects reliably and adaptively. Although significant advances have been made in gripper-based robotic grasping~\cite{gripper_0, gripper_1, gripper_2, gripper_3, gripper_4, gripper_5}, the limited complexity of gripper structures restricts their adaptability to objects with intricate geometries.

Dexterous grasping~\cite{DexGraspNet, UniDexGrasp, UniDexGrasp++, dexterous_grasping_0, dexterous_grasping_1, dexterous_grasping_2, dexterous_grasping_3, dexterous_grasping_4, dexterous_grasping_5} introduces advanced multi-fingered manipulation, enabling more versatile grasps for objects of diverse shapes. However, controlling the highly dexterous multi-fingered system poses significant challenges for traditional analytical techniques~\cite{DexGraspNet, dexterous_control_0, dexterous_control_1, dexterous_control_2}. Recent advances have utilized learning-based methods to enable effective dexterous manipulation. One 
approach decomposes the grasping process into two stages: generating a static grasp pose and then performing a dynamic grasping through trajectory planning or goal-conditioned reinforcement learning~\cite{UniDexGrasp, ILAD, dexterous_grasping_6, grasp_generation_0, grasp_generation_1, grasp_generation_2, grasp_generation_3, grasp_generation_4}.  Although promising diversity, the generated static grasp poses are often not validated in dynamic settings, which adversely affects the overall success. Alternatively, another line of approach directly learns the entire grasping process through expert demonstrations from humans or reinforcement learning agents~\cite{UniDexGrasp++, DAPG, human_demonstration_0, human_demonstration_1, human_demonstration_2, learn_demonstration_0, learn_demonstration_1, dexterous_grasping_8, dexterous_grasping_9, dexterous_grasping_10, dexterous_grasping_11}. While effective, these approaches either require complex training pipelines or suffer significant performance degradation when a single policy is applied across a broad range of objects, due to the limited number of training objects and expert demonstrations, as well as the constrained capacity of their policy networks.

\begin{figure*}[!t]
    \centering
    \includegraphics[width=0.99\linewidth]{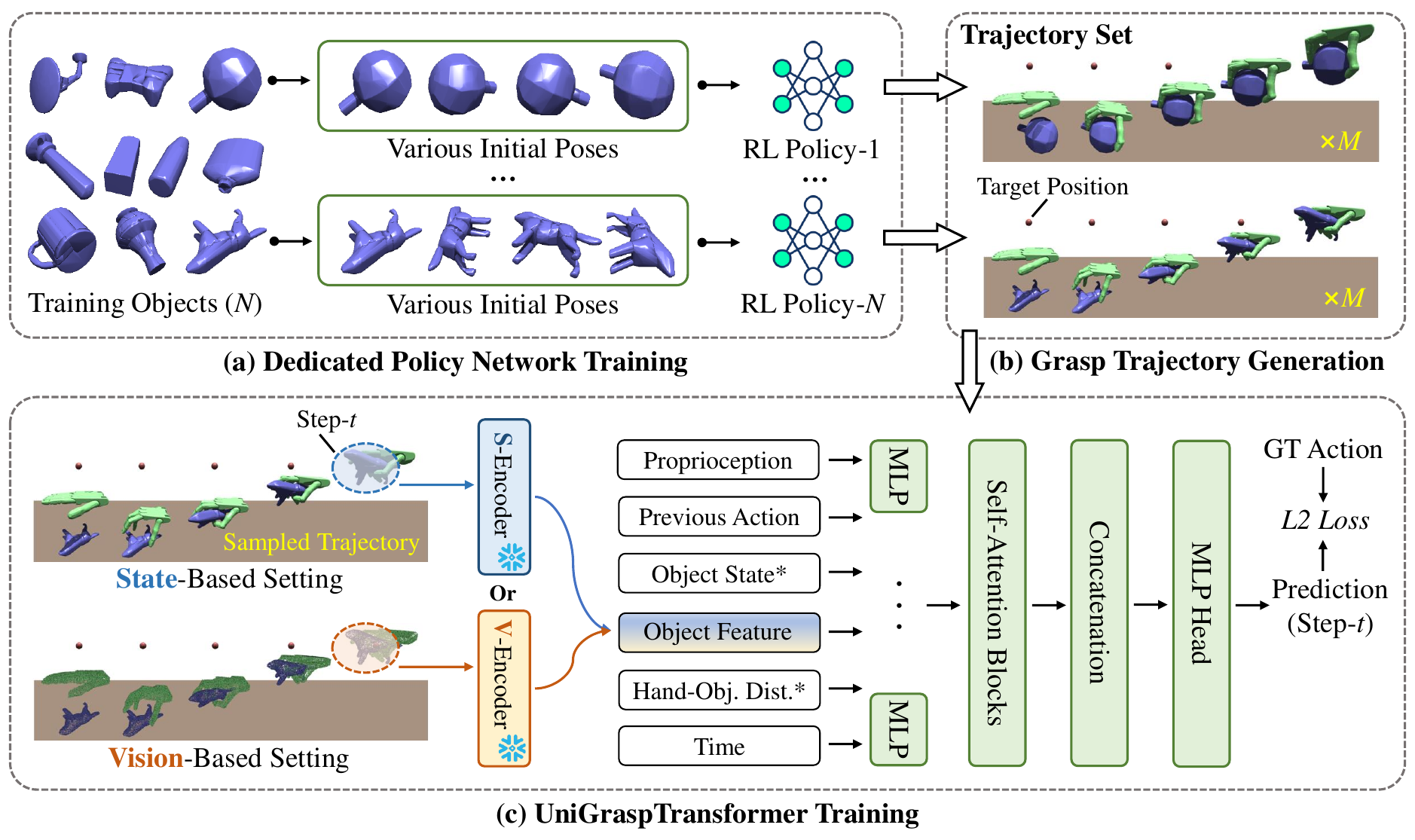}
    \vspace{-2mm}
    \caption{Overview of UniGraspTransformer. (a) Dedicated policy network training: each individual RL policy network is trained to grasp a specific object with various initial poses. (b) Grasp trajectory generation: 
    each policy network generates $M$ successful grasp trajectories, forming a trajectory set $\mathcal{D}$. (c) UniGraspTransformer training: trajectories from $\mathcal{D}$ are used to train UniGraspTransformer, a universal grasp network, in a supervised manner. We investigate two settings—state-based and vision-based—with the primary difference being in the input representation of object state and hand-object distance, as indicated by ``*'' in the figure. The architecture of S-Encoder and V-Encoder can be found in Figure~\ref{fig:autoencoder}.}
    \vspace{-3mm}
    \label{fig:overview}
\end{figure*}

To overcome these limitations, we extend the latter approach by proposing a novel pipeline that integrates online reinforcement learning with large-model offline distillation, simplifying training while improving scalability and grasping performance.

\noindent\textbf{Policy Distillation.}
Policy distillation~\cite{policy_distillation_0, policy_distillation_1, policy_distillation_2, GSL, online_distillation_0, online_distillation_1, imitate_learning_0, imitate_learning_1} provides an effective approach for transferring knowledge from high-performance policies to a single universal policy, promoting both model compactness and generalization across diverse tasks.

In robotics, recent works have focused on combining imitation learning and reinforcement learning~\cite{human_demonstration_0, human_demonstration_1, human_demonstration_2, imitate_learning_0, imitate_learning_1, UniDexGrasp++} to enable student agents to learn from teacher demonstrations. This research generally follows two main approaches based on the source of demonstrations. The first approach directly trains the student policy on pre-collected human demonstrations, such as teleoperated human motions or recorded human videos~\cite{human_demonstration_0, human_demonstration_1, human_demonstration_2, human_demonstration_3}. While effective, gathering extensive demonstrations can be costly, particularly for complex tasks like dexterous grasping with diverse objects in varied poses, limiting the student's generalization capacity. The second approach generates demonstrations using pre-trained policies within the generalist-specialist learning framework~\cite{GSL, progressive_distillation_0, progressive_distillation_1, progressive_distillation_2, UniDexGrasp, UniDexGrasp++, learn_demonstration_1}. Here, the task space is divided into sub-tasks, with reinforcement learning policies specialized and trained for each. These policies are then distilled into a single universal policy, enhancing the agent’s ability to generalize across the full task space. Despite the progress made, a single network handling a broad range of objects often experiences performance drops due to the limited teacher policies and the constrained capacity of student networks, which struggle to capture the complexity of the entire task space.

Our approach addresses these challenges by initially training dedicated policies (i.e., teachers) for each object, resulting in a dataset of 3,200 objects and 3.2 million grasping trajectories. We then leverage UniGraspTransformer (i.e., student) to perform offline universal policy learning, utilizing up to 12 self-attention blocks to process diverse grasping trajectories and better preserve the diversity of teacher policies. This approach significantly enhances dexterous grasping performance across various settings.
\section{Methodology}
\label{sec:method}

\noindent\textbf{Problem Formulation.} 
The objective is to train a robust \textit{universal network}, UniGraspTransformer, that enables a dexterous, five-fingered robotic hand to grasp a variety of tabletop objects in diverse initial poses. Isaac Gym 3.0~\cite{IsaacGym} is utilized as our simulator.

\begin{table*}[!t]
    \centering
    \begin{tabular}{p{4cm}|p{12.5cm}}
        \toprule
Input Type      & Elements (Dimension) \\
\midrule
Proprioception (167) & Wrist position (3) and rotation (3); Finger-joint angle (22), angular velocity (22) and force (22); Finger-tip position (5$\times$3), quaternion rotation (5$\times$4), linear velocity (5$\times$3), angular 
velocity (5$\times$3), force (5$\times$3) and torque (5$\times$3). \\
\midrule
Previous Action (24) & Wrist force (3) and torque (3); Finger-joint angles (18).  \\
\midrule 
Object State (16) & Object center (3), quaternion rotation (4), linear velocity (3), and angular velocity (3); Object-goal distance (3). \\
\midrule
Hand-Object Distance (36) & Hand body points to object point cloud distances (36). \\
\midrule
 Time (29) & Current time (1); Sine-cosine time embedding (28).\\
        \bottomrule
    \end{tabular}
    \vspace{-2mm}
    \caption{Input types for our dedicated policy networks, organized into five groups. Definitions for each element within these groups are provided in the appendix. These inputs are also applicable to our UniGraspTransformer.}
    \label{tab:RL-input}
    \vspace{-3mm}
\end{table*}

\noindent\textbf{Dexterous Hand.} In our implementation, we use the Shadow Hand~\cite{ShadowHand}, which has 18 active degrees of freedom (DOFs) in the fingers—5 for the thumb, 4 for the little finger, and 3 each for the remaining fingers—along with an additional 6 DOFs at the wrist. This gives the dexterous hand a total of 24 active DOFs. The wrist’s active DOFs are controlled via force and torque, while the fingers’ active DOFs are managed through joint angles. In addition, each finger, except for the thumb, has a passive DOF that won't be directly controlled.

\noindent\textbf{Overview.} As shown in Figure~\ref{fig:overview}, the UniGraspTransformer training process comprises three main stages: 1) \textit{Dedicated Policy Network Training} (Section~\ref{sec:policy}), where individual reinforcement learning (RL) policy networks are trained, each dedicated to a single object across various initial poses; 2) \textit{Grasp Trajectory Generation} (Section~\ref{sec:traj-gen}), where each policy network generates $M$ successful grasping trajectories for downstream training. Each trajectory is a sequence of steps capturing the comprehensive knowledge of the environment, including robotic action (e.g., finger-joint angles) and object state (e.g., pose and point cloud); 3) \textit{UniGraspTransformer Training}, where all successful grasping trajectories from various objects and initial poses are used to train UniGraspTransformer in both state-based (Section~\ref{sec:UniGraspTransformer}) and vision-based (Section~\ref{sec:vision-based}) settings. This supervised training process allows UniGraspTransformer to generalize well to both seen and unseen objects.

\subsection{Dedicated Policy Network Training}
\label{sec:policy}
Our training set consists of 3,200 unique tabletop objects. For each object, we train a dedicated policy network across various initial poses, using PPO~\cite{PPO} as our reinforcement learning optimization algorithm. During training, each object is randomly rotated to enhance the initial pose diversity. Once trained, each policy network can successfully grasp its corresponding object across a range of poses.

\noindent\textbf{Input.} Table~\ref{tab:RL-input} summarizes the input types for our dedicated policy networks, organized into five groups: 167-d proprioception, 24-d previous action, 16-d object state, 36-d hand-object distance, and 29-d time. These groups are concatenated into a single 272-d input vector.

\noindent\textbf{Network Architecture.} Each policy network is a 4-layer MLP with hidden dimensions of \{1024, 1024, 512, 512\}, followed by an action prediction head implemented as a single fully connected layer. This head outputs a 24-d vector (18 DOFs for the fingers and 6 DOFs for the wrist) representing the action for the current time step. The value network shares the same architecture as the policy network, also comprising 4 MLP layers with identical hidden dimensions, but it outputs a single scalar value.

\noindent\textbf{Reward Function.} 
The reward function $R$ is defined as:
\begin{equation}
\label{eq:reward}
R = R_{d} + (1-f_{c})R_{o} + f_{c}(R_{l} + R_{g} + R_{s}),
\end{equation}
where the grasp reward $R_{d}$ penalizes the distance between the dexterous hand and the object, encouraging the hand to maintain contact with the object surface for a secure grasp; the contact flag $f_{c}$ is set to 1 if the distance between the hand and the object is below a specified threshold; the reward $R_{o}$ encourages the hand to remain open until contact is made with the object; once contact is established (i.e., $f_{c}=1$), the lift reward $R_{l}$ encourages the hand to perform the lifting action; the goal reward $R_{g}$ penalizes the distance between the object and the target goal; and the success reward $R_{s}$ provides a bonus when the object successfully reaches the goal. The formal definitions of all rewards are available in the appendix. The reward function in Eq.~\ref{eq:reward} is applied to each grasp trajectory, consisting of $T$ steps. In our implementation, we set $T=200$.

\subsection{Grasp Trajectory Generation}
\label{sec:traj-gen}

Each dedicated policy network is now able to grasp its assigned object across various initial poses, achieving an average success rate of 94.1\% across the 3,200 training objects. For each object, we randomly rotate it and use its corresponding policy network to generate a successful grasp trajectory. This process is repeated $M$ ($M=1000$) times per object, resulting in a dataset $\mathcal{D}$ of $3,200 \times M$ successful grasp trajectories. Each trajectory, $\mathcal{T}=\{(S_1,A_1),\dots, (S_t,A_t), \dots, (S_T,A_T)\}$, is a sequence of steps, where $A_t$ represents robotic action (18 active DOFs for the fingers and 6 active DOFs for the wrist) at timestep-$t$, and $S_t$ represents the observation of proprioception, previous action, object state, hand-object distance, time, and object point cloud, as defined in Table~\ref{tab:RL-input}. The dataset $\mathcal{D}$ is then used to train our UniGraspTransformer in a supervised manner, as described in Section~\ref{sec:UniGraspTransformer}.

\subsection{UniGraspTransformer Training}
\label{sec:UniGraspTransformer}
The objective is to use the generated trajectory dataset $\mathcal{D}$ to train a \textit{universal} grasp network, UniGraspTransformer, capable of grasping a variety of tabletop objects in diverse initial poses. UniGraspTransformer is designed to generalize to both seen objects from the training set and previously unseen objects within either seen or unseen categories.

\noindent\textbf{Settings.} We train UniGraspTransformer under two settings: (1) a state-based setting, where object point clouds are perfectly accurate, with direct access to object’s positions and rotations, and (2) a vision-based setting, where object point clouds are estimated and reconstructed using five cameras mounted at the top and borders of the table, with object's positions and rotations estimated rather than directly obtained. The primary distinction between these two settings is the method of acquiring object point clouds and the availability of oracle-level object states. 

We use the state-based setting to illustrate the key components of UniGraspTransformer, and describe its adaptation to the vision-based setting in Section~\ref{sec:vision-based}.

\begin{figure}[!t]
    \centering
    \includegraphics[width=0.8\linewidth]{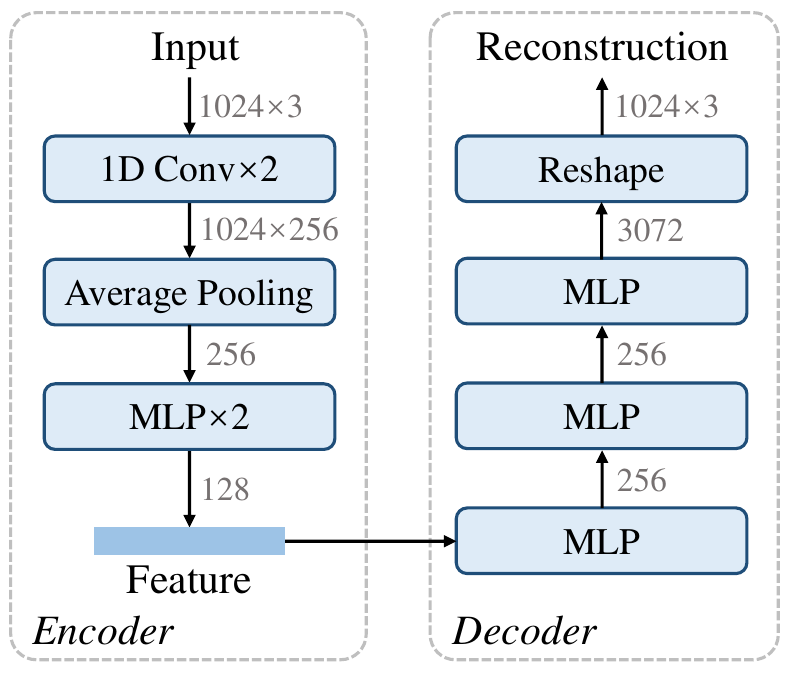}
    \vspace{-2mm}
    \caption{
    Illustration of the network architecture of the object point cloud encoder, S-Encoder, in the state-based setting. The process begins with sampling 1,024 points from the object point cloud, producing an input with a dimension of $1024 \times 3$. This input is passed through the \textit{encoder}, producing a 128-dimensional object feature, which the \textit{decoder} then uses to reconstruct the 1,024 sampled points, with the Chamfer Distance serving as the loss function. During inference, only the \textit{encoder} is used to convert an object point cloud into a 128-dimensional object feature.
    }
    \vspace{-3mm}
    \label{fig:autoencoder}
\end{figure}

\noindent\textbf{Input Process.} As detailed in Section~\ref{sec:traj-gen}, each trajectory $\mathcal{T}=\{(S_1,A_1),\dots, (S_t,A_t), \dots, (S_T,A_T)\}$ consists of $T$ time steps. At each step $t$, we train UniGraspTransformer with $S_t$—which includes information on proprioception, previous action, object state, hand-object distance, time, and object point cloud—as input to predict the corresponding action $A_t$, a 24-d vector. Table~\ref{tab:RL-input} provides dimensions for each component: proprioception (167-d), previous action (24-d), object state (16-d), hand-object distance (36-d), and time (29-d).
For encoding the object point cloud, an object encoder named S-Encoder, which has a similar structure as the PointNet~\cite{PointNet}, is trained specifically for the state-based setting, encoding the point cloud into an 128-d feature (see Figure~\ref{fig:autoencoder}). Thus, the model has six input vectors: 167-d proprioception, 24-d previous action, 16-d object state, 26-d hand-object distance, 29-d time and 128-d object feature. As illustrated in Figure~\ref{fig:overview}(c), each input vector is mapped to a 256-d token via an individual MLP network. These six tokens serve as the input to UniGraspTransformer.

\noindent\textbf{Network Architecture and Loss Function.} As illustrated in Figure~\ref{fig:overview}(c), UniGraspTransformer consists primarily of several self-attention blocks, followed by a 4-layer MLP head that outputs a 24-d action prediction. By default, we use 12 self-attention blocks. For a given data pair $(S_t, A_t)$ at time step $t$ from trajectory $\mathcal{T}$, we first convert $S_t$ into six tokens as previously described. These tokens are then fed into UniGraspTransformer to produce the prediction $P_t$. The model is optimized using L2 loss, defined as $\mathcal{L}=||A_t - P_t||_2$.

\begin{table*}[!t]
%\small
%\setlength\tabcolsep{5pt}
    \centering
    % \resizebox{\linewidth}{!}{
    \begin{tabular}{l|ccc|ccc}
        \toprule
        \multirow{3}{*}{Method} & \multicolumn{3}{c|}{State-Based Setting (\%)} & \multicolumn{3}{c}{Vision-Based Setting (\%)} \\  \cmidrule(lr){2-4} \cmidrule(lr){5-7}
         & \multirow{2}{*}{Seen Obj.} & Unseen Obj.  & Unseen Obj. & \multirow{2}{*}{Seen Obj.} & Unseen Obj. & Unseen Obj. \\
         &  & Seen Cat.  & Unseen Cat. &  & Seen Cat. & Unseen Cat. \\
        \midrule
        PPO$^\dagger$~\cite{PPO} & 24.3 & 20.9 & 17.2 & 20.6 & 17.2 & 15.0 \\
        DAPG$^\dagger$~\cite{DAPG} & 20.8 & 15.3 & 11.1 & 17.9 & 15.2 & 13.9 \\
        ILAD$^\dagger$~\cite{ILAD} & 31.9 & 26.4 & 23.1 & 27.6 & 23.2 & 20.0 \\
        GSL$^\dagger$~\cite{GSL} & 57.3 & 54.1 & 50.9 & 54.1 & 50.2 & 44.8 \\
        UniDexGrasp~\cite{UniDexGrasp} & 79.4 & 74.3 & 70.8 & 73.7  & 68.6 &  65.1 \\
        UniDexGrasp++~\cite{UniDexGrasp++} & 87.9  & 84.3  & 83.1  &  85.4 &  79.6 & 76.7\\
        \midrule
        Ours & \textbf{91.2}  & \textbf{89.2}  & \textbf{88.3}  &  \textbf{88.9} &  \textbf{87.3} & \textbf{86.8} \\
        \bottomrule
    \end{tabular}
    \vspace{-2mm}
    \caption{Comparison with state-of-the-art methods using a universal model for dexterous robotic grasping across both state-based and vision-based settings, evaluated by success rate. Evaluation on unseen objects from either seen or unseen categories assesses the models' generalization capability. $\dagger$ indicates results reported in UniDexGrasp++~\cite{UniDexGrasp++}. ``Obj.'': Objects. ``Cat.'': categories.}
    \label{tab:compare_sota}
    \vspace{-3mm}
\end{table*}

\subsection{Adaptation to the Vision-Based Setting}
\label{sec:vision-based}
\noindent\textbf{Input Adaptation.} In the vision-based setting, we use five cameras mounted at the table’s top and borders to estimate the object point clouds. The estimated point clouds consist of two components: 1) the partial object point cloud and 2) the hand point cloud, which is segmented and removed in our implementation. In the state-based setting, the object point clouds are uniformly sampled from the object mesh, which are complete and accurate. This difference affects the input to UniGraspTransformer as follows: 1) for \textit{object state} representation, we use the center of the partial object point cloud as the object position and apply PCA on this partial cloud to represent the object rotation; 2) we use the partial object point cloud to calculate \textit{hand-object distance}; and 3) we re-train a dedicated \textit{object encoder}, termed V-Encoder, to extract features from the partial object points. Other configurations, such as network architecture, loss function, and supervision signals, remain unchanged.

\noindent\textbf{V-Encoder.} In Section~\ref{sec:UniGraspTransformer}, for the state-based setting, we train an S-Encoder (see Figure~\ref{fig:autoencoder}) that encodes complete object point clouds into object features. In contrast, in the vision-based setting, we only have access to partial object point clouds. To extract their features, we re-train a V-Encoder, maintaining the same network architecture as the S-Encoder. The key modifications are as follows: 1) the input consists of 1,024 sampled points from the partial object point cloud; 2) a distillation loss is applied to regularize the V-Encoder’s latent features, with supervision provided by the latent features of the corresponding complete object point cloud extracted by the S-Encoder. After training, the V-Encoder can extract a 128-d object feature from partial object point clouds.

\section{Experiments}
\label{sec:exp}

\noindent\textbf{Datasets.}
We utilize the UniDexGrasp++~\cite{UniDexGrasp++} dexterous grasping dataset, comprising 3,200 objects across 133 categories for training. Evaluation is conducted on these 3,200 seen objects, as well as on 140 unseen objects from seen categories and 100 unseen objects from unseen categories. For seen objects, we generate test initial poses separately from those used during training, applying this protocol for both dedicated policy network evaluation and UniGraspTransformer evaluation.

\noindent\textbf{Evaluation Protocols.} Following UniDexGrasp++~\cite{UniDexGrasp++}, each object is randomly rotated and dropped onto the table to enhance the diversity of its initial poses. This process is repeated 1,000 times for robust evaluation. A grasp is considered successful if the object reaches the target goal within $T=200$ steps. We report the average success rate across all objects and grasp attempts. The evaluation is performed in two configurations: a state-based setting and a vision-based setting. The state-based setting represents an ideal scenario where the object point cloud is entirely accessible and flawlessly accurate. Conversely, in the vision-based setting (see Section~\ref{sec:vision-based}), the object point cloud is reconstructed using depth images captured from five different viewpoints by five cameras. Additionally, the dexterous hand may partially occlude the object, resulting in only a portion of the object’s point cloud being accessible.

\noindent\textbf{Implementation Details.} For a single dedicated policy network, we create 1,000 simulation environments, and update the policy network every 16 steps over 10K iterations, with a learning rate of 3e-4. Training is conducted parallelly on 16 NVIDIA V100 GPUs, requiring 80 hours for 3,200 dedicated policies. The UniGraspTransformer model is trained with a batch size of 800 trajectories over 100 epochs, using a fixed learning rate of 1e-4. Training is performed on 8 NVIDIA A100 GPUs and takes around 70 hours to complete. The object encoders, both state-based and vision-based, are trained on point cloud data with a batch size of 100. Training runs for 800K iterations with a learning rate of 5e-4 on an NVIDIA A100 GPU, requiring 40 hours.

\subsection{Main Results}
\textbf{Dedicated Policy Networks.} As detailed in Section~\ref{sec:policy}, we train an individual policy network for each of the 3,200 training objects. During evaluation, each policy is used exclusively with its corresponding object, achieving an average success rate of 94.1\%. However, these dedicated policies cannot be evaluated on unseen objects, as they lack generalization capability.

\noindent\textbf{UniGraspTransformer.} Table~\ref{tab:compare_sota} compares our UniGraspTransformer with state-of-the-art methods using a universal model for dexterous robotic grasping across both state-based and vision-based settings. Our model outperforms UniDexGrasp++~\cite{UniDexGrasp++} by 3.3\% and 3.5\% on seen objects in the state-based and vision-based settings, achieving success rates of 91.2\% and 88.9\%, respectively. Furthermore, our UniGraspTransformer demonstrates strong generalization capabilities, achieving high success rates on unseen objects and unseen categories in both settings. It surpasses UniDexGrasp++~\cite{UniDexGrasp++} by 4.9\% (7.7\%) and 5.2\% (10.1\%) on unseen objects from seen categories and unseen objects from unseen categories in the state-based (vision-based) setting. The transition from seen to unseen categories further verifies the generalization capability of our approach, with only a minimal drop in success rate observed in both the state-based (91.2\% to 88.3\%) and vision-based (88.9\% to 86.8\%) settings.

\subsection{Ablation Studies}
Unless otherwise specified, the ablation studies are conducted on the 3,200 seen objects.

\begin{table}[!t]
    \centering
    \begin{tabular}{l|ccc}
        \toprule
Trajectory Number ($M$)  & 0.2K & 0.5K & 1K \\
\midrule
Success Rate (\%) & 87.2 & 89.3 & \textbf{91.2} \\
        \bottomrule
    \end{tabular}
    \vspace{-2mm}
    \caption{For each of the 3,200 dedicated policy networks, we generate $M$ successful grasping trajectories, which are then distilled into UniGraspTransformer. We analyze the impact of varying $M$.}
    \label{tab:ablation_tra_num}
    \vspace{-1mm}
\end{table}

\begin{table}[!t]
    \centering
    \begin{tabular}{l|ccc}
        \toprule
Self-Attention Blocks ($K$) & - & 6 & 12 \\
\midrule
Success Rate (\%) & 85.5 & 89.7 & \textbf{91.2} \\
        \bottomrule
    \end{tabular}
    \vspace{-2mm}
    \caption{Analysis of the impact of different numbers of self-attention blocks in UniGraspTransformer. The model consists of $K$ self-attention blocks followed by a 4-layer MLP head. ``-'' indicates a configuration that uses only the MLP head without any self-attention blocks.}
    \label{tab:ablation_att_blocks}
    \vspace{-1mm}
\end{table}

\begin{table}[!t]
    \centering
    \begin{tabular}{l|cccc}
        \toprule
Object Number & 400 & 800 & 1,600 & 3,200 (All)\\
\midrule
Success Rate (\%) & 92.5 & 91.8 & 91.3 & \textbf{91.2} \\
        \bottomrule
    \end{tabular}
    \vspace{-2mm}
    \caption{Analysis of distilling varying numbers of dedicated policy networks (one policy per object) into UniGraspTransformer. The evaluation is conducted on the corresponding seen object set.}
    \label{tab:ablation_obj_num}
    \vspace{-1mm}
\end{table}

\begin{table}[!t]
    \centering
    \begin{tabular}{l|cc}
        \toprule
Method  & DAgger & UniGraspTransformer \\
 % Method & (Online) & (Offline) \\
\midrule
Success Rate (\%) & 88.2 & \textbf{92.5} \\
        \bottomrule
    \end{tabular}
    \vspace{-2mm}
    \caption{Analysis of distilling 400 dedicated policies into one universal policy, via online or offline methods. The evaluation is conducted on the corresponding seen object set.}
    \label{tab:ablation_distill}
    \vspace{-4mm}
\end{table}

\noindent\textbf{Scalability of UniGraspTransformer.} Our approach employs a two-step process: initially, dedicated policy networks are individually trained for each object using reinforcement learning to generate a set of successful grasp trajectories. These trajectories are then distilled into a single universal network, UniGraspTransformer. This design allows us to investigate the scalability of UniGraspTransformer in terms of trajectory set size, network capacity, and the number of objects it can manage. The following studies are conducted under the state-based setting.
\begin{itemize}
    \item \textit{Trajectory Set Size.} As shown in Table~\ref{tab:ablation_tra_num}, UniGraspTransformer demonstrates strong scalability in handling an increasing number of trajectories. The success rate improves as the number of grasping trajectories per object used for training grows, indicating the model's ability to effectively learn and generalize from diverse trajectories across various objects.
    \item \textit{Network Capacity.} As illustrated in Figure~\ref{fig:overview}(c), our UniGraspTransformer consists of input MLP layers, $K$ self-attention blocks, and a 4-layer MLP head. Table~\ref{tab:ablation_att_blocks} presents an analysis of the impact of network capacity on performance. The results show that increasing the number of self-attention blocks enhances the success rate, indicating that UniGraspTransformer scales effectively with additional self-attention layers. Specifically, the success rate improves from 85.5\% (without self-attention blocks) to 89.7\% with 6 blocks and finally reaches 91.2\% with 12 blocks. 
    \item \textit{Object Number.} Table~\ref{tab:ablation_obj_num} shows the success rate of our UniGraspTransformer when distilling varying numbers of object-specific policy networks, evaluated across different object sets with 400, 800, 1,600, and 3,200 objects. The results indicate a high and stable success rate, with a slight decline as the number of objects increases.
    \item \textit{Online vs. Offline.} Table~\ref{tab:ablation_distill} compares the online distillation method: DAgger with MLPs, to our offline distillation method: UniGraspTransformer. The results highlight the advantages of offline distillation with large policy networks when handling a diverse set of teacher policies.
\end{itemize}

\noindent\textbf{Input of UniGraspTransformer.}
Table~\ref{tab:ablation_input} presents an analysis of different input components and their effects on performance. The basic input includes proprioception, previous actions, and object state. We progressively enhance the input by incorporating features from the state-based object encoder, hand-object distances, and temporal information. The performance improves consistently, indicating that UniGraspTransformer effectively utilizes diverse information sources to enhance robotic grasping capabilities.

\begin{table}[!t]
    \centering
    \resizebox{0.46\textwidth}{!}{\begin{tabular}{cccccc|c}
        \toprule
Proprio- & Prev.  & Obj.  & Obj.  & Hand- & \multirow{2}{*}{Time} & SR \\ 
ception  & Action & State & Feat. & Obj. Dist.  & & (\%)  \\ \midrule
        % \checkmark & & &  & &  \\
        % \checkmark & \checkmark &  &  &  &  & \\
        \checkmark & \checkmark & \checkmark &  &  &  & 78.4 \\
        \checkmark & \checkmark & \checkmark & \checkmark &  &  & 86.6 \\
        \checkmark & \checkmark & \checkmark & \checkmark & \checkmark & & 89.9 \\
        \checkmark & \checkmark & \checkmark & \checkmark & \checkmark & \checkmark & \textbf{91.2} \\
        \bottomrule
    \end{tabular}}
    \vspace{-2mm}
    \caption{Effects of different input components on UniGraspTransformer Training. ``Prev.'': previous. ``Obj.'': Object. ``Feat.'': feature. ``Dist.'': distance. ``SR'': success rate.}
    \label{tab:ablation_input}
    \vspace{-1mm}
\end{table}

\begin{table}[!t]
    \centering
    \begin{tabular}{cc|c}
        \toprule
Center of & PCA of & Success Rate  \\
Partial Object & Partial Object & (\%) \\
\midrule
 & & 83.2 \\
\checkmark & & 86.4 \\
\checkmark & \checkmark & \textbf{88.9} \\
        \bottomrule
    \end{tabular}
    \vspace{-2mm}
    \caption{Utilizing approximate estimations—specifically, the center and PCA of partial object point clouds—enhances performance compared to the baseline without these estimations.}
    \label{tab:ablation_vision-input}
    \vspace{-5mm}
\end{table}

\noindent\textbf{UniGraspTransformer in the Vision-Based Setting.} As described in Section~\ref{sec:vision-based}, the vision-based setting involves estimating object point clouds from five cameras, resulting in incomplete point clouds. Unlike the state-based setting—where complete object states, including rotation, are accessible—object rotation information is unavailable in this configuration. A baseline solution in the vision-based setting is to omit the point cloud center and object rotation inputs. Alternatively, the center of partial object point clouds can substitute the center of full point clouds, and principal component analysis (PCA) can approximate object rotations. Table~\ref{tab:ablation_vision-input} compares these alternatives with the baseline, demonstrating that these estimations improve performance over the baseline lacking any estimation.

In Section~\ref{sec:vision-based}, we introduce a distillation loss for training the vision-based object encoder, enabling it to extract a 128-dimensional feature from partial object point clouds. This distillation process transfers knowledge from the state-based object encoder, which encodes complete object point clouds into a single feature, to the vision-based encoder. Table~\ref{tab:ablation_vision-AE} studies the impact of using a vision-based object encoder with and
without distillation loss on UniGraspTransformer’s performance, showing a 2.2\% improvement in success rate, underscoring the value of distillation in training the vision-based encoder.

\begin{table}[!t]
    \centering
    \begin{tabular}{c|c}
        \toprule
   Distillation & Success Rate (\%) \\
\midrule
%Full Object & & 86.3 \\
%Full Object & \checkmark & 88.5 \\
 & 86.7 \\
 \checkmark & \textbf{88.9} \\

        \bottomrule
    \end{tabular}
    \vspace{-2mm}
    \caption{Impact of using a vision-based object encoder with and without distillation loss on UniGraspTransformer's performance.}
    \vspace{-1mm}
    \label{tab:ablation_vision-AE}
    %\vspace{-3mm}
\end{table}

\begin{table}[!t]
    \centering
    \begin{tabular}{c|cc|c}
        \toprule
$R_{o}$ & $R_{d}$ w/ Point Cloud &  $R_{d} $ w/ Center &  SR  (\%)\\
\midrule
 &  & \checkmark & 90.3 \\
 & \checkmark & & 92.9 \\
 \checkmark & \checkmark & &  \textbf{94.1} \\
        \bottomrule
    \end{tabular}
    \vspace{-2mm}
    \caption{Impact of incorporating reward $R_{o}$ and two variants of reward $R_{d}$ on the performance of dedicated policy networks.}
    \label{tab:ablation_reward}
    \vspace{-4mm}
\end{table}

\noindent\textbf{Reward Functions for Dedicated Policy Network.} As outlined in Section~\ref{sec:policy}, the dedicated policies are trained with reward functions defined in Eq.\ref{eq:reward}. We explore two reward functions: (1) the grasp reward $R_{d}$, which penalizes the distance between the dexterous hand and the object, and (2) the reward $R_{o}$, which encourages the hand to stay open until it contacts the object. In our default setup, $R_{d}$ is computed by measuring the distances between 36 hand points and the object point clouds. In Table~\ref{tab:ablation_reward}, we examine an alternative version where $R_{d}$ is based on distances between the 36 hand points and the object center. Additionally, Table~\ref{tab:ablation_reward} assesses the impact of including or excluding the reward $R_{o}$. A well-designed reward function enhances the performance of all dedicated policy networks, achieving an average success rate of 94.1\% across 3,200 training objects. Please refer to our supplementary material for more details.

\begin{figure}[!t]
    \centering
    \includegraphics[width=0.90\linewidth]{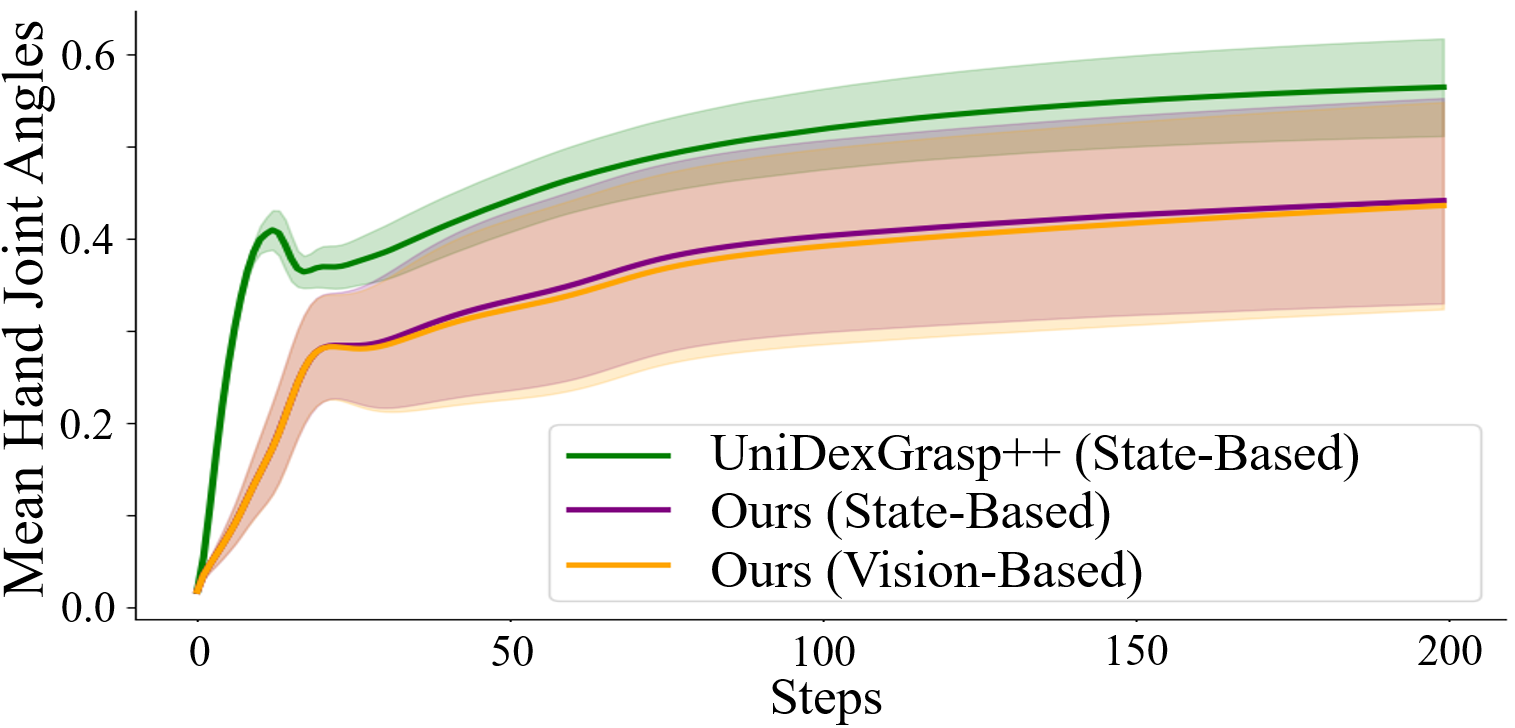}
    \vspace{-2mm}
    \caption{Quantitative analysis of grasp pose diversity.}
    \vspace{-2mm}
    \label{fig:diversity_curve}
\end{figure}

\begin{figure}[!t]
    \centering
    \includegraphics[width=0.95\linewidth]{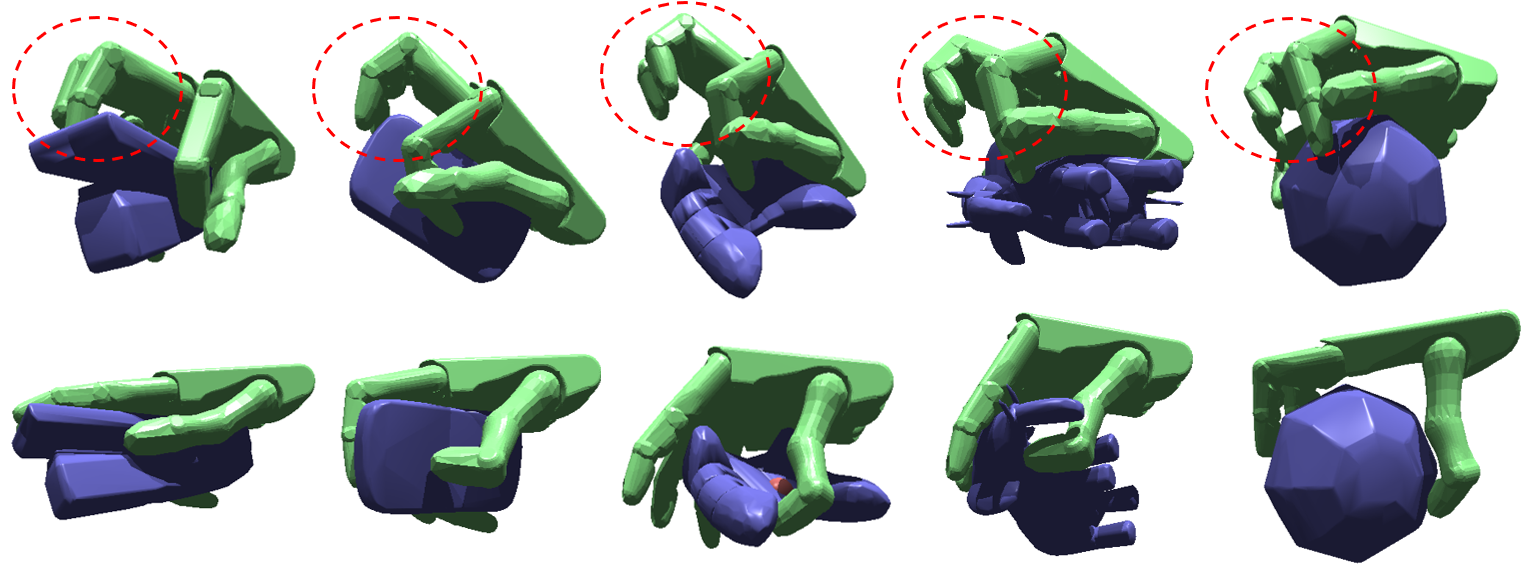}
    \vspace{-2mm}
    \caption{Comparison of grasp poses generated by the state-based universal model from the UniDexGrasp++~\cite{UniDexGrasp++} (top row) and our UniGraspTransformer (bottom row). Each column displays two distinct grasp poses for the same object with the same initial pose.}
    \vspace{-4mm}
    \label{fig:diversity_pose}
\end{figure}

\noindent\textbf{Diversity Analysis on Grasp Pose.} 
Figure~\ref{fig:diversity_curve} provides a quantitative comparison of grasp pose diversity between the state-based UniDexGrasp++~\cite{UniDexGrasp++} and both the state-based and vision-based versions of UniGraspTransformer. During inference, each model generates 10 successful 200-step trajectories for each of the 3,200 training objects, with mean hand joint angles (normalized) used to represent the hand state at each step. The plotted range in Figure~\ref{fig:diversity_curve} demonstrates that UniGraspTransformer exhibits a broader range, indicating its capability to produce diverse grasp poses across a variety of objects. Figure~\ref{fig:diversity_pose} presents visual examples that highlight the greater diversity of grasp poses generated by our model compared to the previous method.

\section{Conclusion}
\label{sec:conclusion}

In this work, we introduce UniGraspTransformer, a universal Transformer-based network that streamlines the training process for dexterous robotic grasping while enhancing scalability, flexibility, and diversity in grasping strategies. Our approach simplifies traditional complex pipelines by employing dedicated reinforcement learning-based policy networks for individual objects, followed by an efficient offline distillation process that consolidates successful grasping trajectories into a single, scalable model. Our UniGraspTransformer is capable of handling thousands of objects in varied poses, exhibiting robustness and adaptability across both state-based and vision-based settings. Notably, our model significantly improves grasp success rates on seen, unseen within-category, and fully novel objects, outperforming the current state-of-the-art with substantial gains in success rates across various settings. 

{
    \small
    \bibliographystyle{ieeenat_fullname}
    \bibliography{main}
}

\appendix
\clearpage
\maketitlesupplementary
% \appendix

\section{Implementation Details}
\label{sec:Method_Supp}

\begin{figure}[!t]
    \centering
    \includegraphics[width=0.9\linewidth]{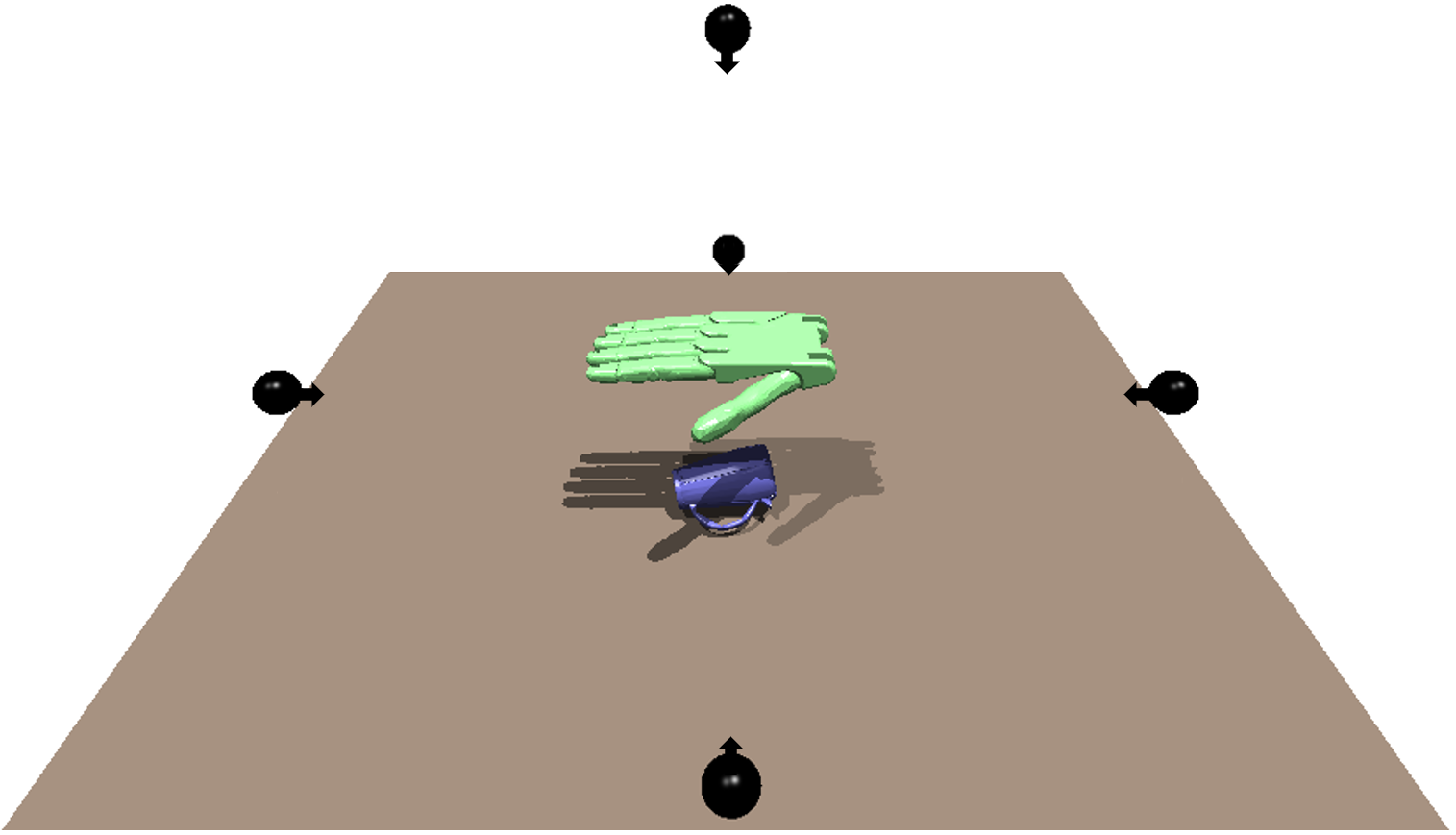}
    \vspace{-2mm}
    \caption{Illustration of the simulation environment.}
    \vspace{-2mm}
    \label{fig:env}
\end{figure}

\subsection{Environment Setup}

\noindent\textbf{Initialization.}
We use Isaac Gym 3.0~\cite{IsaacGym} to build simulation environments, each containing a table (brown), an object placed on top (blue), a controllable Shadow Hand (green)~\cite{ShadowHand}, and five surrounding cameras (black), as illustrated in Figure~\ref{fig:env}. The system’s origin is defined at the center of the table, where all objects are initially placed. The Shadow Hand is positioned 0.2 meters above the table center, with the goal located 0.3 meters above the table center.

For each object utilized in our project, we randomly drop it onto the table with arbitrary rotations to generate a dataset comprising 12K static tabletop poses. This dataset is divided into three subsets for specific purposes: 10K poses for dedicated policy training, 1K poses for offline trajectory generation, and 1K poses for evaluation.

\noindent\textbf{Task Definition.} The objective is to develop a robust universal policy capable of controlling the Shadow Hand~\cite{ShadowHand} to grasp and transport a diverse range of tabletop objects to a designated midair goal position. Each grasping consists of 200 execution steps and is deemed successful if the positional difference between the object and the goal remains within a predefined threshold by the end of the sequence.

\begin{figure}[!t]
    \centering
    \includegraphics[width=0.9\linewidth]{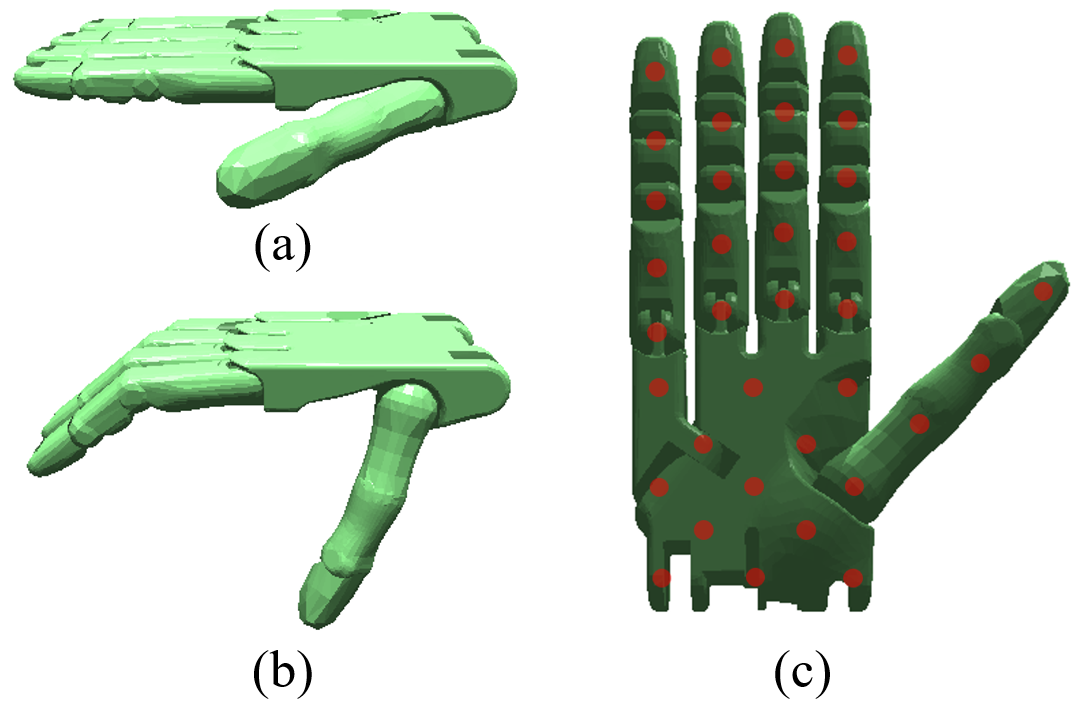}
    \vspace{-2mm}
    \caption{Shadow Hand poses. (a) Initial pose at the first frame. (b) Pre-contact opening pose used in dedicated policy training. (c) 36 selected hand points for computing hand to object distance.}
    \vspace{-2mm}
    \label{fig:poses}
\end{figure}

\noindent\textbf{Observation Space.}
At each simulation step, the observation space of state-based UniGraspTransformer includes a 167-d proprioceptive state of the hand, a 24-d representation of the hand’s previous action, a 16-d object state, a 128-d object visual feature, a 36-d hand-to-object distance, and a 29-d time embedding, as detailed in Table 1 of the main paper. During dedicated RL policy training, the 128-d object visual feature is excluded to enhance training efficiency. For vision-based UniGraspTransformer training, the original 16-d object state is replaced by the center position of the partial object point cloud (3-d) and its three principal component axes ($3\times3$-d). Additionally, we compute 36 distances between 36 selected points on the Shadow Hand and the partial object point cloud, as illustrated in Figure~\ref{fig:poses}(c).

\noindent\textbf{Action Space.}
The action space comprises motor commands for 24 actuators on the Shadow Hand. The first 6 actuators manage the wrist's position and orientation through applied forces and torques, while the remaining 18 actuators control the positions of the finger joints. The action values are normalized to a range of -1 to 1 according to the specifications of the actuators.

\noindent\textbf{Camera Setup.} Following a similar approach to UniDexGrasp++~\cite{UniDexGrasp++}, five RGBD cameras are mounted around the table, as illustrated in Figure~\ref{fig:env}. The cameras are positioned relative to the table center at coordinates (0.0, 0.0, 0.55), (0.5, 0.0, 0.15), (-0.5, 0.0, 0.15), (0.0, 0.5, 0.15), and (0.0, -0.5, 0.15), with their focal points aligned at (0, 0, 0.15). In the vision-based setting, the depth images captured by these cameras are fused to generate a scene point cloud, from which the partial point cloud of the object is segmented.

\subsection{Dedicated Policy Training}
\noindent\textbf{PPO.} Proximal Policy Optimization~\cite{PPO} is a widely used model-free, on-policy reinforcement learning algorithm that simultaneously learns a policy and estimates a value function. We utilize PPO to train dedicated RL policies for each of the 3,200 objects. Both the policy and value networks are implemented as 4-layer MLPs with hidden dimensions of \{1024, 1024, 512, 512\}. At each simulation step, the policy network takes the current observation as input and outputs a 24-d action, while the value network predicts a 1-d value. The simulation environment then executes the action, updates the observation, and calculates the corresponding reward. The policy and value networks are updated every 16 simulation steps using the collected observations, actions, values, and rewards. Each dedicated RL policy is trained on an NVIDIA V100 GPU for a total of 10,000 update iterations, taking approximately 3 hours to complete.

\noindent\textbf{Reward Function.}
The reward function described in Eq.(1) of the main paper comprises five components: $R_d$, $R_o$, $R_l$, $R_g$, and $R_s$. These reward components are governed by a contact flag $f_c$, which indicates whether the hand is in contact with the object.

The distance reward $R_d$ penalizes the average Chamfer Distance between the hand points $H_i$ and the object point cloud $P_{obj}$, promoting contact and encouraging the hand to maintain a secure grasp on the object's surface. Specifically, 36 points $\{H_{i}\}_{i=1}^{36}$ are selected on the Shadow Hand for this calculation, as illustrated in Figure~\ref{fig:poses}(c).
\begin{equation}
\label{eq:R_d}
R_{d} = -\omega_d \frac{1}{36}\sum_{i=1}^{36} ChamferDistance(H_i, P_{obj}),
\end{equation}
where the reward weight $\omega_d$ is set to 1.0.

The contact flag $f_{c}$ is set to 1 if the average Chamfer Distance between the hand points and the object point cloud falls below a predefined threshold $\lambda_c=0.06$. This is determined as follows:
\begin{equation}
\label{eq:f_c}
f_c = \mathbbm{1}[\frac{1}{36}\sum_{i=1}^{36} ChamferDistance(H_i, P_{obj}) < \lambda_c],
\end{equation}
where $\mathbbm{1}[\cdot]$ denotes the indicator function.

Inspired by DexGraspNet~\cite{DexGraspNet}, before the contact is established, the opening reward $R_o$ penalizes deviations of the current hand pose $q$ from a predefined opening pose $q_{open}$, as depicted in Figure~\ref{fig:poses}(b). This encourages the hand to remain open until it makes contact with the object. The reward is calculated as:
\begin{equation}
\label{eq:R_o}
R_o = -\omega_o \parallel q - q_{open} \parallel_2,
\end{equation}
where the reward weight $\omega_o$ is set to 0.1.

Once contact is established, the rewards $R_l$, $R_g$, and $R_s$ are introduced to guide the grasping process:
\begin{itemize}
    \item Lift reward ($R_l$): This reward encourages the hand to perform a lifting action $a_{z}$ along the z axis:
    \begin{equation}
\label{eq:R_l}
R_l = \omega_l (1 + a_{z}),
\end{equation}
where $\omega_l$ is set to 0.1.
\item Goal reward ($R_{g}$): This reward penalizes the Euclidean distance between the object center position $x_{obj}$ and the target goal position $x_{goal}$:
\begin{equation}
\label{eq:R_g}
R_g = -\omega_g \parallel x_{obj} - x_{goal} \parallel_2,
\end{equation}
where $\omega_g$ is set to 2.0.
\item Success reward ($R_{s}$): This reward provides a bonus when the object successfully reaches the goal position, defined by a threshold $\lambda_g=0.05$:
\begin{equation}
\label{eq:R_s}
R_s = \omega_s \mathbbm{1}[\parallel x_{obj} - x_{goal} \parallel_2 < \lambda_g],
\end{equation}
where $\omega_s$ is set to 1.0.
\end{itemize}

%The reward weights are set as $\omega_d=1.0$, $\omega_o=0.1$, $\omega_l=0.1$, $\omega_g=2.0$, and $\omega_s=1.0$ respectively.

\subsection{Grasp Trajectory Generation}
Our 3,200 dedicated RL policies achieve an average success rate of 94.1\% across all 3,200 training objects. For each object, we randomly initialize it in diverse poses and apply its corresponding RL policy to generate $M=1000$ successful trajectories, which are used for offline training of the UniGraspTransformer model. Each trajectory, $\mathcal{T}=\{(S_1,A_1),\dots, (S_t,A_t), \dots, (S_T,A_T)\}$, consists of a sequence of steps. Here, $A_t$ represents robotic action at timestep-$t$, and $S_t$ captures the environment state, including proprioception (167-d), previous action (24-d), object state (16-d), hand-object distance (36-d), and time embedding (29-d), as detailed in Table 1 of the main paper. Additionally, we save the complete object point cloud ($1024\times3$-d) and the partial object point cloud ($1024\times3$-d), which are used to generate object features to train the state-based and vision-based versions of the UniGraspTransformer model.

\subsection{Point Cloud Encoder Training}

\noindent\textbf{S-Encoder.}
To train our S-Encoder, we use a dataset consisting of 3,200 point clouds of seen objects, denoted as $\{P_{i}\}_{i=1}^{3200}$, where each $P_{i}$ represents the canonical point cloud of a specific object. During each training iteration, a batch of 100 object point clouds is randomly sampled from this dataset. For each point cloud in the batch, indexed by $j$ ($j=1,2,\dots,100$), the centroid $\mathbf{c}_{j}$ is subtracted to center the point cloud, followed by the application of a random rotation matrix $R_{j}$. The resulting transformed point cloud is expressed as $\hat{P}_{j} = R_{j} \left( P_{j} - \mathbf{c}_{j} \right)$, which serves as input to the S-Encoder.

The S-Encoder, as part of an encoder-decoder framework~\cite{PointNet}, processes $\hat{P}_{j}$ to produce a latent feature $z_{j}$. This latent feature is then passed to the decoder, which reconstructs the point cloud, yielding $\tilde{P}_{j}$. The model is trained by minimizing the reconstruction loss $\mathcal{L}_{CD}$, defined as the Chamfer Distance between the original transformed point cloud $\hat{P}_{j}$ and its reconstruction $\tilde{P}_{j}$:
\begin{equation}
\label{eq:l_ae_S}
\mathcal{L}_{CD} = ChamferDistance(\hat{P}_{j}, \tilde{P}_{j}).
\end{equation}
The S-Encoder is trained for 800K iterations on an NVIDIA A100 GPU. After training, the state-based object features are generated by encoding the complete object point clouds using the trained S-Encoder.

%we use a dataset of 3,200 seen object point clouds, denoted as $\{PC_{i}\}_{i=1}^{3200}$, where each $PC_{i}$ represents the canonical point cloud of a specific object. In each training iteration, we randomly sample a batch of 100 object point clouds from this dataset. Each object point cloud in the batch, indexed by $j$ (where $j=1,2,…,100$), is first centered by subtracting its centroid $\mathbf{c}_{j}$, after which a random rotation $R_{j}$ is applied. The final transformed point cloud, denoted as $\hat{PC}_{j} = R_{j} \left( PC_{j} - \mathbf{c}_{j} \right)$ served as the input to our S-Encoder. 

%The S-Encoder is part of an encoder-decoder. The S-Encoder processes the transformed point cloud $\hat{PC}_{j}$ and generates a latent feature $z_{j}$. This feature $z_{j}$ is then passed to the decoder $\mathcal{D_{S}}$, which reconstructs the point cloud, producing $\tilde{PC}_{j}$. The model is trained by minimizing a reconstruction loss $\mathcal{L}_{CD}$, defined as the Chamfer Distance between the original transformed point cloud $\hat{PC}_{j}$ and its reconstruction $\tilde{PC}_{j}$.

%The S-Encoder is trained on an NVIDIA A100 GPU over 800K iterations, after which, the state-based object features are encoded using the saved complete object point clouds.

\begin{figure}[!t]
    \centering
    \includegraphics[width=0.95\linewidth]{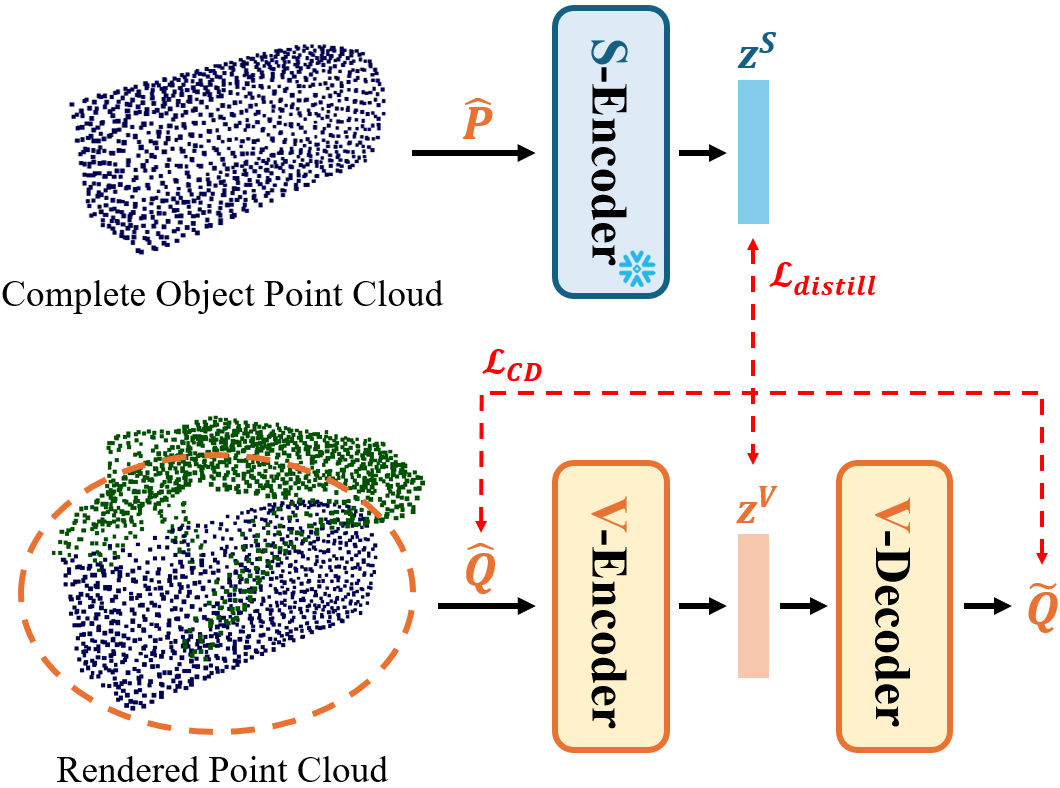}
    \vspace{-2mm}
    \caption{V-Encoder training with distillation.}
    \vspace{-2mm}
    \label{fig:distill_encoder}
\end{figure}

\noindent\textbf{V-Encoder.}
The V-Encoder is trained using a knowledge distillation approach, leveraging the pre-trained S-Encoder and the grasp trajectories $\mathcal{T}$ generated by the dedicated RL policies, as illustrated in Figure~\ref{fig:distill_encoder}. In each training iteration, a batch of 100 steps is randomly sampled from the generated trajectories. Both the complete object point cloud $P_{t}$ and the partial point cloud $Q_{t}$ are centered by subtracting their mean positions. The centered complete point cloud $\hat{P}_{t}$ is passed through the pre-trained S-Encoder (with frozen weights), producing a latent feature $z_{t}^{S}$. Simultaneously, the centered partial point cloud $\hat{Q}_{t}$ is fed to the V-Encoder, which outputs both a latent feature $z_{t}^{V}$ and a reconstructed point cloud $\tilde{Q}_{t}$. 

The V-Encoder is optimized using two loss functions:
\begin{itemize}
    \item Feature Distillation loss ($\mathcal{L}_{distill}$): This L2 loss measures the difference between the latent features produced by the S-Encoder and the V-Encoder:
    \begin{equation}
\label{eq:l_distill}
\mathcal{L}_{distill} = \parallel z_{t}^{S} - z_{t}^{V}\parallel_2
\end{equation}
\item Reconstruction loss ($\mathcal{L}_{CD}$): This is the Chamfer Distance between the centered partial point cloud $\hat{Q}_{t}$ and its reconstruction $\tilde{Q}_{t}$: \begin{equation}
\label{eq:l_ae_V}
\mathcal{L}_{CD} = ChamferDistance(\hat{Q}_{t}, \tilde{Q}_{t}),
\end{equation}
\end{itemize}

The total loss for training the V-Encoder is defined as:
\begin{equation}
\label{eq:l_sum_V}
\mathcal{L} = \omega_{CD}\mathcal{L}_{CD} + \omega_{distill}\mathcal{L}_{distill},
\end{equation}
where the weights are set to $\omega_{CD}=1.0$ and $\omega_{distill}=0.1$. The V-Encoder is trained on an NVIDIA A100 GPU for 800K iterations. After training, the vision-based object features are generated by encoding the partial object point clouds using the trained V-Encoder.

\subsection{UniGraspTransformer Training}

\noindent\textbf{Input Types.} The UniGraspTransformer is trained using the generated grasp trajectories and encoded object features under two configurations, as outlined in Table~\ref{tab:universalpolicy-input}:
\begin{itemize}
    \item State-Based Setting: The complete object point clouds are assumed to be perfectly accurate and are encoded using the S-Encoder. Object states, including positions, rotations, and velocities, are directly accessible, as detailed in Table 1 of the main paper.
    \item Vision-Based Setting: Partial object point clouds are reconstructed and segmented from depth data captured by five cameras mounted above and around the table. These object features are encoded using the V-Encoder, and object states are estimated rather than directly accessed.
\end{itemize}
The key differences between the inputs for the state-based and vision-based UniGraspTransformer are:
\begin{itemize}
    \item For the object state representation, the vision-based setting uses the center of the partial object point cloud (3-d) as the object position and three principal component axes (9-d) to represent object orientation.
    \item The object feature is derived from the partial object point cloud and encoded using the V-Encoder in the vision-based setting.
    \item The hand-object distance is computed using the partial object point cloud in the vision-based setting.
\end{itemize}

%We train our UniGraspTransformer using the generated grasp trajectories and encoded object features under two settings as shown in Table~\ref{tab:universalpolicy-input}: (1) a state-based setting, where the complete object point clouds are perfectly accurate and encoded using S-Encoder, with direct access to object states, including positions, rotations, and velocities as listed in Table~1 of the main paper, and (2) a vision-based setting, where the partial object point clouds are reconstructed and segmented using five cameras mounted at the top and borders of the table, with object features encoded using V-Encoder and object states estimated. 

%Specifically, there are three main differences between inputs of the state-based and vision-based UniGraspTransformer: 1) for \textit{object state*} representation, we use center of the partial object point cloud (3-d) as the object position and use three PCA axes (9-d) to represent the object rotation; 2) we use V-Encoder to encode the partial object point cloud into the \textit{object feature*}; and 3) we use the partial object point cloud to calculate the \textit{hand-object distance*}.

\begin{table}[!t]
    \centering
    \begin{tabular}{c|c}
        \toprule
\multicolumn{2}{c}{Input of UniGraspTransformer} \\
\midrule
State-Based & Vision-Based  \\
\midrule
Proprioception (167) & Proprioception (167) \\
Previous Action (24) & Previous Action (24) \\
Object State (16) & Object State* (12) \\
Object Feature (128) & Object Feature* (128) \\
Hand-Obj. Dist. (36) & Hand-Obj. Dist.* (36) \\
Time (29) &Time (29) \\
        \bottomrule
    \end{tabular}
    \vspace{-2mm}
    \caption{Input types for state-based and vision-based UniGraspTransformer, organized into six groups.}
    \label{tab:universalpolicy-input}
    \vspace{-3mm}
\end{table}

\noindent\textbf{Training Process.}
Each trajectory step consists of six observation groups, as detailed in Table~\ref{tab:universalpolicy-input}, paired with a ground truth action $A_t$. The UniGraspTransformer processes these observations as follows: (1) The six observation groups are converted into six 256-dimensional tokens using individual single-layer MLPs; (2) These tokens are passed through 12 self-attention layers~\cite{Attention}, producing six refined 256-dimensional features; (3) The six features are concatenated into a single 1536-dimensional representation, which is then processed by a 4-layer MLP to predict the final 24-d action $P_t$. The model is optimized using a single L2 loss, defined as: $\mathcal{L}=||A_t - P_t||_2$.

Training is conducted on a dataset of 3,200 objects with 3.2 million trajectories, using a batch size of 800 trajectories (each with 200 steps) over 100 epochs. The process is carried out on 8 NVIDIA A100 GPUs and takes approximately 70 hours to complete. The average L2 loss at convergence is around 0.015.

%Each trajectory step sample includes six observation groups as listed in Table~\ref{tab:universalpolicy-input}, along with a ground truth action $A_t$. Our UniGraspTransformer first converts these six observation groups into six 256-d tokens via six single-layer MLPs. These six tokens are then passed through 12 self-attention layers, generating six refined 256-d features. These six 256-d features are then concatenated to form a single 1536-d representation, which is passed through a 4-layer MLP to predict the final 24-d action $P_t$. UniGraspTransformer is optimized using a single L2 loss, defined as $\mathcal{L}=||A_t - P_t||_2$.

%Our UniGraspTransformer is trained on a dataset of 3.2 million trajectories, and a batch size of 800 trajectories (each contains 200 steps) over a totoal of 100 epochs. Training is performed on 8 NVIDIA A100 GPUs and takes around 70 hours to complete. The average L2 loss at the convergance is around 0.015.

\section{Experiment Details}
\label{sec:Experiment_Supp}

\subsection{Baseline Methods}
The implementation of baseline methods listed in Table~2 of the main paper is outlined below. Additional details can be found in UniDexGrasp++\cite{UniDexGrasp++}.

\noindent\textbf{PPO.} This reinforcement learning baseline directly trains a state-based universal model using PPO with all training objects. The vision-based universal policy is derived from the state-based policy through distillation using DAgger~\cite{online_distillation_0}.

\noindent\textbf{DAPG.} Demo Augmented Policy Gradient (DAPG)~\cite{DAPG} is a widely used imitation learning method that leverages expert demonstrations to reduce RL sampling complexity. In this baseline, grasp trajectories generated via motion planning serve as demonstrations to train a state-based deep RL policy. The vision-based universal policy is then distilled from the state-based policy using DAgger~\cite{online_distillation_0}.

\noindent\textbf{ILAD.} ILAD~\cite{ILAD} enhances the generalization capabilities of DAPG~\cite{DAPG} by introducing an imitation learning objective focused on the object's geometric representation. In this baseline, a pipeline similar to DAPG~\cite{DAPG} is implemented.

\begin{figure}[!t]
    \centering
    \includegraphics[width=0.99\linewidth]{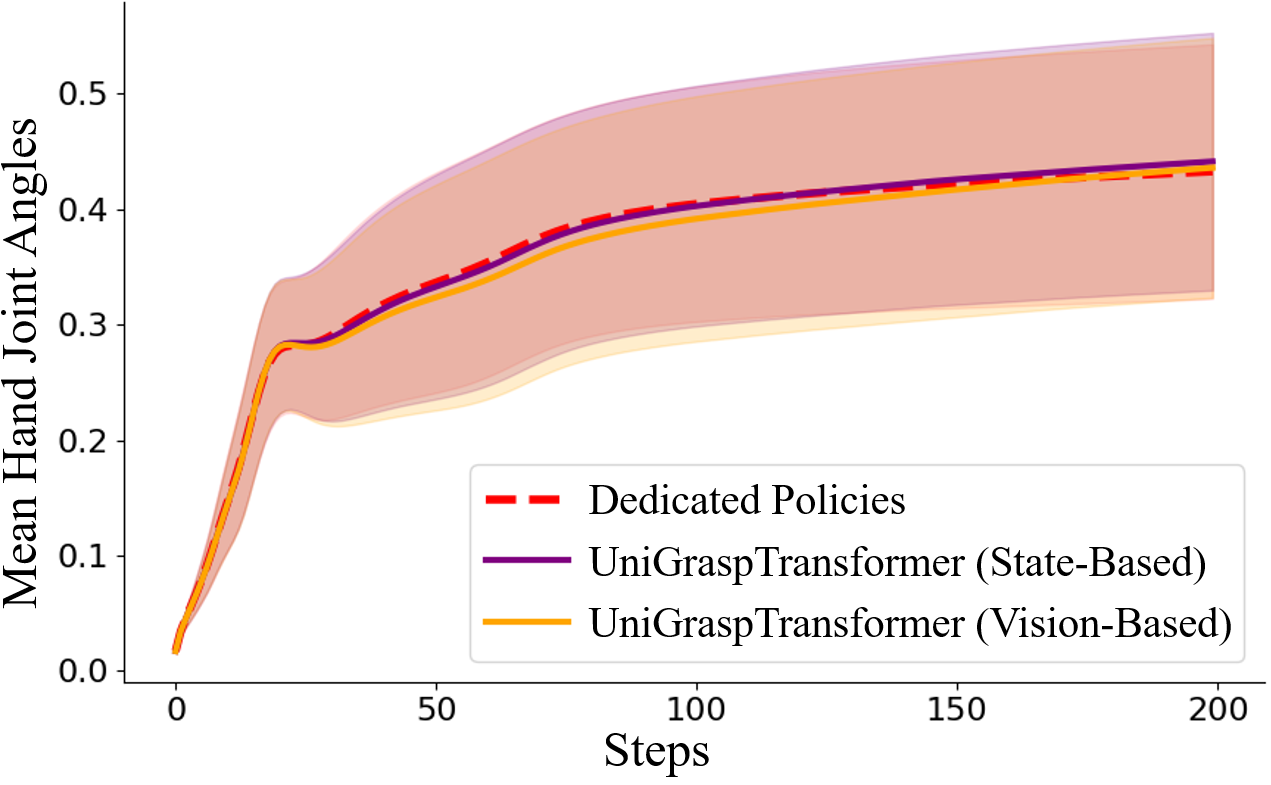}
    \vspace{-2mm}
    \caption{Quantitative analysis of grasp pose diversity.}
    \vspace{-2mm}
    \label{fig:decay}
\end{figure}

\noindent\textbf{GSL.} Generalist-Specialist Learning (GSL)~\cite{GSL} begins by training a generalist policy using PPO over the entire task space. Specialists are then fine-tuned to master each subset of the task space. The final generalist is trained using DAPG~\cite{DAPG}, leveraging demonstrations generated by the trained specialists.

\noindent\textbf{UniDexGrasp.} UniDexGrasp~\cite{UniDexGrasp} decomposes the grasping task into two stages: static grasp pose generation followed by dynamic grasp execution via goal-conditioned reinforcement learning. First, an IPDF-based~\cite{IPDF} grasp pose generator is trained using all training objects. An Object Curriculum Learning protocol is then applied, starting reinforcement learning with a single object and gradually incorporating similar objects to train a state-based universal policy. Finally, DAgger~\cite{online_distillation_0} is used to distill the state-based universal policy into a vision-based universal policy.

\noindent\textbf{UniDexGrasp++.} UniDexGrasp++~\cite{UniDexGrasp++} builds on the Generalist-Specialist Learning framework by integrating geometry-based clustering during specialist training, where each specialist focuses on a group of geometrically similar objects. Additionally, it introduces a generalist-specialist iterative process in which specialists are repeatedly trained from the generalist, followed by generalist distillation.

\begin{figure}[!t]
    \centering
    \includegraphics[width=1\linewidth]{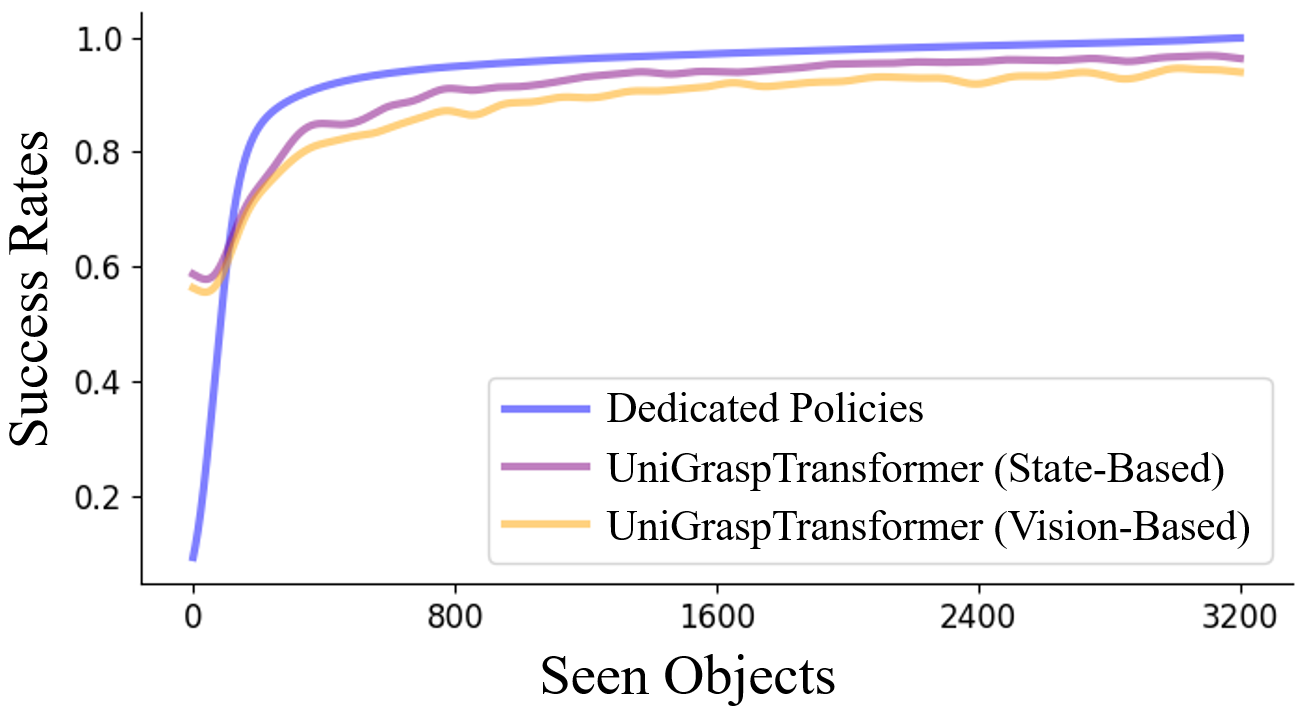}
    \vspace{-2mm}
    \caption{Success rates across seen objects.}
    \vspace{-2mm}
    \label{fig:success}
\end{figure}

% \begin{figure*}[!t]
%     \centering
%     \includegraphics[width=0.95\linewidth]{figures/diversity_compare.png}
%     % \vspace{-2mm}
%     \caption{Comparison of grasp poses. (a) UniDexGrasp++ often grasps different objects using similar poses, characterized by twisted and floated middle fingers. (b) In contrast, UniGraspTransformer adapts to objects of various shapes, exhibiting a diverse array of grasp poses.}
%     % \vspace{-2mm}
%     \label{fig:diversity_compare}
% \end{figure*}

\section{More Analysis}
\noindent\textbf{From Dedicated to Universal.} Our 3,200 dedicated RL policies achieve an average success rate of 94.1\% across all 3,200 training objects. In comparison, the UniGraspTransformer achieves success rates of 91.2\% (88.9\%) on 3,200 seen objects, 89.2\% (87.3\%) on 140 unseen objects from seen categories, and 88.3\% (86.8\%) on 100 unseen objects from unseen categories under the state-based (vision-based) settings, respectively.

As depicted in Figure~\ref{fig:decay}, the UniGraspTransformer effectively replicates the grasping trajectories generated by the dedicated RL policies through offline distillation. While there is a minor performance drop from 94.1\% to 91.2\% (88.9\%) in the state-based (vision-based) setting, as illustrated in Figure~\ref{fig:success}, the model demonstrates robust generalization and efficiency.

%\noindent\textbf{Performance Drop due to Offline Distillation}
%As shown in Figure~\ref{fig:decay}, our UniGraspTransformer, utilizing the offline distillation, successfully replicates the grasping trajectories generated from the dedicated RL teachers, with slight performance drop from 94.1\% to 91.2\% (88.9\%) in the state-based (vision-based) setting, as shown in Figure~\ref{fig:success}.

\begin{figure*}[!t]
    \centering
    \includegraphics[width=0.92\linewidth]{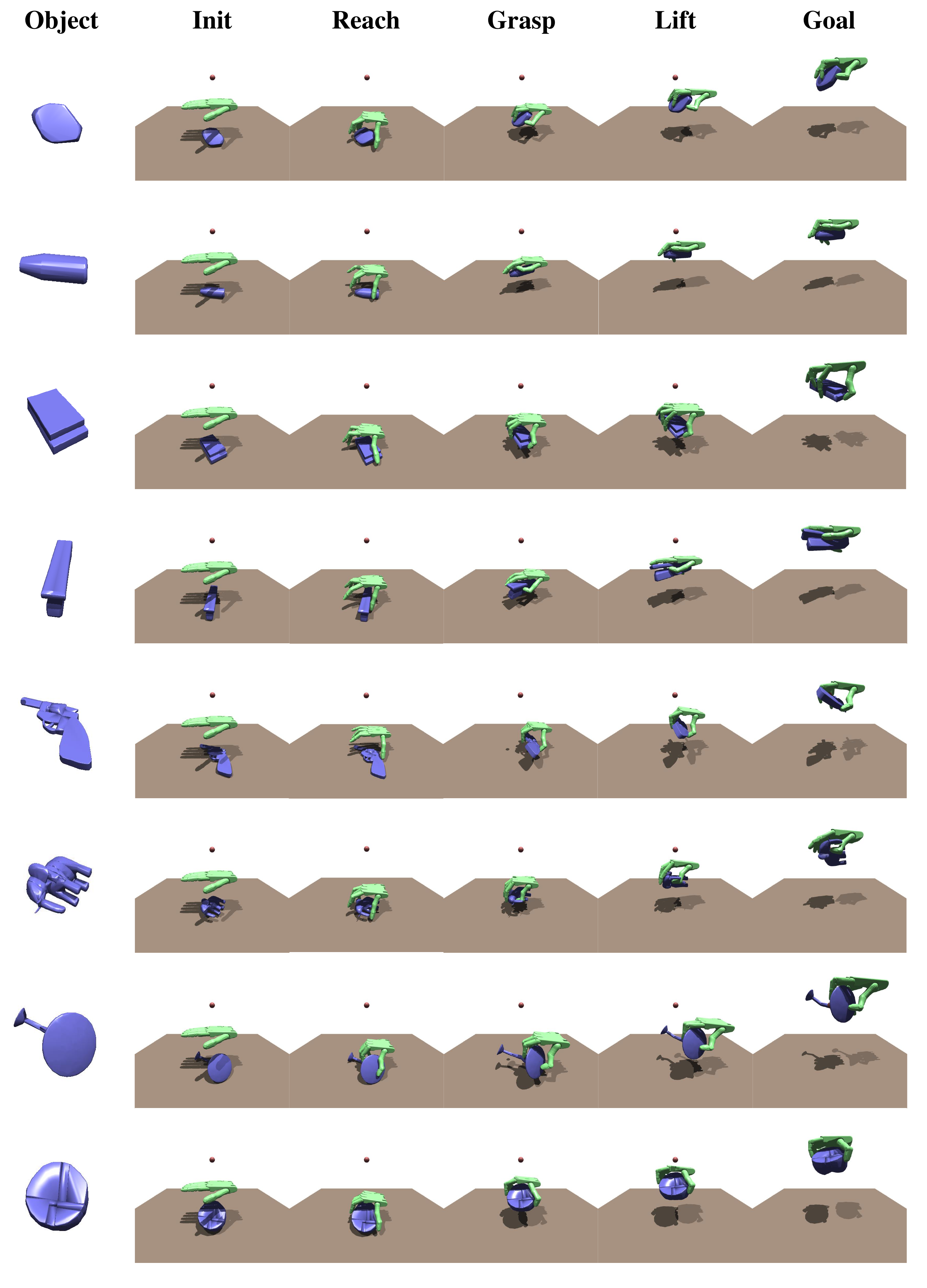}
    \vspace{-4mm}
    \caption{Qualitative analysis of the grasp pose diversity achieved by UniGraspTransformer.}
    \vspace{-2mm}
    \label{fig:demo}
\end{figure*}

\noindent\textbf{Qualitative Results.}
The progressive online distillation approach~\cite{GSL} employed in UniDexGrasp++\cite{UniDexGrasp++} results in a universal policy that tends to grasp different objects using similar poses. In contrast, our UniGraspTransformer, utilizing a larger model and an offline distillation framework, demonstrates the ability to grasp objects of various shapes with a wide range of diverse poses. This increased diversity in grasping strategies is further highlighted in Figure~\ref{fig:demo}.

\noindent\textbf{Real-World Deployment.} We extend the deployment of our vision-based UniGraspTransformer to a real-world environment using the Inspire Hand~\cite{InspireHand}, which features six active DoFs for its fingers. The training process remains identical to that used for the Shadow Hand. Demonstration videos showcasing grasping across 12 distinct objects are provided in the supplementary materials.

\end{document}